\documentclass[10pt,twocolumn,letterpaper]{article}

\usepackage{cvpr}
\usepackage{times}
\usepackage{epsfig}
\usepackage{graphicx}
\usepackage{amsmath}
\usepackage{amssymb}

\usepackage{booktabs}

\usepackage[pagebackref=true,breaklinks=true,letterpaper=true,colorlinks,bookmarks=false]{hyperref}
\cvprfinalcopy 

\def\cvprPaperID{3256} 

\ifcvprfinal\fi
\begin{document}

\title{DeepSkeleton: Skeleton Map for 3D Human Pose Regression}

\author{Qingfu Wan\\
Fudan University\\
{\tt\small qfwan13@fudan.edu.cn}
\and
Wei Zhang\\
Fudan University\\
{\tt\small weizh@fudan.edu.cn}
\and
Xiangyang Xue\\
Fudan University\\
{\tt\small xyxue@fudan.edu.cn}
\\
}

\maketitle

\begin{abstract}
   Despite recent success on 2D human pose estimation, 3D human pose estimation still remains an open problem. A key challenge
   is the ill-posed depth ambiguity nature. This paper presents a novel intermediate
   feature representation named skeleton map for regression. It distills structural context from irrelavant properties of RGB image $\emph{\eg}$ illumination and texture. It is simple, clean and can be easily generated via deconvolution network. For the first 
   time, we show that training regression network from skeleton map alone is capable of meeting 
   the performance of state-of-the-art 3D human pose estimation works. We further exploit the power of multiple 3D hypothesis
   generation to obtain reasonbale 3D pose in consistent with 2D pose detection.
   The effectiveness of our approach is validated on challenging in-the-wild       
   dataset MPII and indoor dataset Human3.6M.

\end{abstract}

\section{Introduction}

A prevalant family of human pose estimation works generally fall into two groups: 2D human pose estimation and 3D human pose estimation. Last year has witnessed the revolution of 2D human pose estimation, thanks to the development of heatmap-based network and the availability of deep residual network \cite{he2016deep}. However, the research on 3D human pose estimation has been significantly lagging behind its counterpart. 

Prior works on 3D human pose estimation can be roughly categorized into two families: \emph{regression methods} and \emph{reconstruction methods}. \emph{Regression methods} directly learn a mapping function from input image to the target 3D joint locations\cite{li20143d, li2014heterogeneous, li2015maximum, moreno20163d, park20163d, sun2017compositional, tekin2016structured, zhou2017towards, tekin2017learning, mehta2016monocular}. As popularized in 3D human pose estimation, the performance is not as excellent as expected. \emph{Reconstruction methods} typically follow a two-stage schema where in the first stage a 2D pose estimator is employed and then 3D pose is reconstructed aiming to minimize the reprojection error via \emph{optimization} \cite{akhter2015pose, bogo2016keep, ramakrishna2012reconstructing, fan2014pose, zhou2017sparse} or \emph{matching} \cite{jahangiri2017generating, chen20163d}. While \emph{optimization} can generate only one 3D output, recent work of Jahangiri and Yuille \cite{jahangiri2017generating} has highlighted the importance of having multiple hypotheses under the widely known problem of depth ambiguity.

A fundamental challenge limiting previous methods from attacking 3D human pose estimation is the insufficient training data. Most top-performing methods in 3D human pose estimation are restricted in laboratory environment where the appearance variation is far less than outdoor scene. On the other hand, there exists no accurate 3D ground truth for in-the-wild dataset to date. Fusing 2D and 3D data, dates back to at least \cite{yasin2016dual}, therefore has become an emerging trend \cite{sun2017compositional, zhou2017towards}. This naturally brings us to a question, \emph{ for accurate 3D human pose estimation, is mixing different data sources really indispensible?}

\begin{figure}[t]
\begin{center}

   \includegraphics[width=0.5\linewidth]{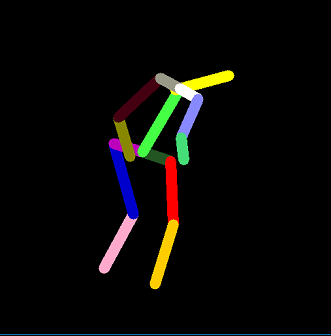}
\end{center}
   \caption{An example of \emph{skeleton map}. 15 body parts are drawn in different colors.}
\label{fig:skeleton}

\end{figure}

In this work, we argue that combined training is not necessary by better exploiting the data that we already have. We put forth a novel expressive intermediate feature representation called \emph{skeleton map}. For an example see Figure ~\ref{fig:skeleton}.
 The core insight is that \emph{human being can easily reason about 3D human pose given skeleton map by our prior knowledge about human kinematics, anatomy and anthropometrics, rather than image observation}. This suggests that most of the image cues, \eg lightning, texture, and human-object interaction are useless for 3D pose inference. 
 
\emph{Skeleton map} is simple, compact and effective. Unlike full human body part segmentation map which requires pixel-wise segmentation, it only models the connection between adjacent joints in the human skeleton. It contains rich structural information and exhibits inherent occlusion-aware property for regression. 

\emph{Skeleton map} is general for both indoor and outdoor scenario. Researchers previously argued that 2D and 3D data complements each other and mixing them is central to better shared feature learning. But with our pure and succinct \emph{skeleton map} representation, we are able to achieve near state-of-the-art performance. We then take a leap forward to generate multiple hypotheses, sharing the same spirit of  ``wisdom of the crowd" \cite{lifshitz2016human}. Different scales of \emph{skeleton maps} are conveniently generated by deconvolutional networks for subsequent regression. It can be intepretered as implicit data augmentation \emph{without data fusion or data synthesis}. This desirable property further resolves the depth ambiguity.

To the best of our knowledge, this is the first work to 
\begin{itemize}
\item[$\bullet$] Perform regression directly from \emph{skeleton map} alone.
\item[$\bullet$] Generate multiple hypotheses \emph{without 3D library}.
\item[$\bullet$] Unite segmentation, pose regression and heatmap detection into a framework we call \emph{DeepSkeleton} for 3D human pose estimation. 
\end{itemize}

\emph{DeepSkeleton} achieves 86.5\emph{mm} average joint error on 3D Human3.6M dataset and delivers considerable 3D poses on 2D MPII dataset.

\section{Related Work}
For a comprehensive review on the literature of 2D human pose estimation we refer the readers to \cite{gong2016human, luvizon2017human, sun2017human}. Here we survey on previous works most relevant to our approach.

\textbf{3D Human Pose Estimation} Li and Chan \cite{li20143d} pioneered the work of direct 3D human pose regression from 2D image. Several approaches have been proposed to learn the pose structure from data.
Tekin \etal \cite{tekin2016structured} learn latent pose representation using an auto-encoder. Zhou \etal \cite{zhou2016deep} integrate a generative forward kinematics layer into the network to learn the joint angle parameters. Realizing the difficulty to minimize per-joint error, \cite{park20163d, mehta2016monocular, sun2017compositional} advocate to predict relative joint locations to multiple joints. More recently, Rogez \etal \cite{rogez2017lcr} simplify the problem by local residual regression in the classified pose class.
 Another research direction has been focused on inferring 3D pose from 2D estimates. The underlying premise is that 2D human pose estimation can be regarded as nearly solved\cite{wei2016convolutional, newell2016stacked, carreira2016human, bulat2016human, chu2016structured, chu2017multi, gkioxari2016chained, lifshitz2016human, yang2016end}. Consequently, the challenge of 3D pose estimation has been shifted from predicting accurate 2D towards predicting depth from RGB image. As an example, Moreno-Noguer \cite{moreno20163d} employs FCN\cite{long2015fully} to obtain pairwise 3D distance matrix from 2D observation for absolute 3D pose recovery. Tekin \etal \cite{tekin2017learning} fuse image space and heatmap space to combine the best of both worlds. Tome \etal \cite{tome2017lifting} adopt an iterative multi-stage architecture where 2D confidence is progressively lifted to 3D and projected back to 2D, ensuring the match between 2D observation and 3D prediction. Zhou \etal \cite{zhou2016sparseness} combine heatmap and 3D geometric prior in EM algorithm to reconstruct 3D skeleton from 2D joints. Zhou \etal \cite{zhou2017towards} output depth from 2D heatmap and intermediate feature maps in a weakly-supervised fashion. Nevertheless, the inherent depth ambiguity presents the foremost obstacle impeding the progress of 3D human pose estimation. Recent techniques ameliorate this issue through dual-source training. Examples include \cite{yasin2016dual, zhou2017towards, sun2017compositional}. Our proposed \emph{DeepSkeleton} stands in contrast to the latent assumption of these works that shared feature learning from both 2D and 3D data is essential. Our results suggest that 2D and 3D data sources may be completely indepedent, which is in line with recent observation \cite{tome2017lifting, martinez2017simple}.
 
\textbf{Exemplar based pose estimation}
Most previous work on 3D human pose estimation rely on generating only one single 3D pose. This is problematic as multiple 3D poses may have similar 2D projection. With early reference to \cite{cho2015accurate}, generating multiple hypotheses has been repurposed for 3D human pose estimation in \cite{jahangiri2017generating}. In their work, a generative model is responsible for generating multiple 3D poses from 3D MoCap library. Similarly, Chen and Ramanan \cite{chen20163d} retrieve 3D pose using neareast neighbour search. \emph{DeepSkeleton} shares some resemblence to these works expect for the need of large-scale offline MoCap dataset.

\textbf{Semantic Segmentation}
Fully convolutional network(FCN)\cite{long2015fully} made the first attempt to solve semantic segmentation using deep neural network. More recent works enjoy the benefits of residual connection \cite{lin2016refinenet, chen2016deeplab}. However, accurate pixel-wise annotation of segmentation map is time-intensive. Different from full human body part segmentation\cite{oliveira2016deep, xia2015zoom, liang2015human} , synthesizing ground truth of our proposed \emph{skeleton map} only requires 2D annotation.

\textbf{Joint Semantic Segmentation and Pose Estimation}
Segmentation and pose estimation are two intrinsically complementary tasks in the sense that pose esmation benefits from the topology of semantic parts, while the estimated pose skeleton offers natural cue for the alignment with part instance. One of the most significant works that jointly solve these two problems is Shotton \etal \cite{shotton2013real}, which derives an intermediate body part segmentation representation for 3D joint prediction from depth image. A large body of literature has been devoted to this field \cite{dong2014towards, ladicky2013human, yamaguchi2012parsing, kohli2008simultaneous, alahari2013pose}. Tripathi \etal \cite{tripathi2017pose2instance} demonstrate the impact of human pose prior on person instance segmentation. Perhaps the most related approach to ours is Xia \etal \cite{xia2017joint}. Their \emph{skeleton label map} serves as a prior to regularize the segmentation task, and is derived from 2D pose prediction. In contrast, our \emph{skeleton map} is independent of 2D detection and is directly taken as input to the subsequent regression module. Another major difference is that they aim to solve multi-person 2D pose estimation, while we target at single-person 3D pose estimation.

\section{Methodology}
Given a RGB image $\mathbf{I} \in {\mathcal{R}}^{224\times224}$ of a person, our goal is to output 3D joint locations $X\in {\mathcal{R}}^{K\times3} $ with joint number $K=16$. We break the problem down into three steps:

\begin{itemize}
\item Segmentation(Section \ref{sec:deconv}): 

For each configuration $p_i=\left\lbrace c_i, l_i \right\rbrace$ with crop scale $c_i$ and stick width $l_i$, \emph{foreground skeleton map} $S_i^f \in {\mathcal{R}}^{56\times56}$ and \emph{background skeleton map} $S_i^b \in {\mathcal{R}}^{56\times56} $ are generated via deconvolutional network ${Deconv}_i$($i=1,...,n$).

\item Regression(Section \ref{sec:regress}): 

$\emph{Skeleton maps}$ $\mathcal{S}=\left\lbrace (S_i^f,S_i^b) |i=1,...n\right\rbrace$ are individually fed into \emph{separate} regression networks $R=\left\lbrace {Regression}_i|i=1,...n\right\rbrace$, where ${Regression}_i$ takes \emph{skeleton map} $S_i^f, S_i^b$ as input and outputs 3D pose hypothesis $X^i$, resulting in multiple 3D hypotheses $\mathcal{H}=\left\lbrace X^i | i=1,...n \right\rbrace $.

\item Matching(Section \ref{sec:matching}): 

To match with 2D observation $x \in \mathcal{R}^{K\times2}$, the hypothesis $X^{\ast} \in \mathcal{H}$ with minimum projection error to 2D joint detection is selected as final output.
\end{itemize}

\subsection{Skeleton Map}
\emph{Skeleton map} $S_i^f$($S_i^g$) draws a stick with width $l_i$ between neighbouring joints in the human skeleton and assigns different colors to distinguish body parts, an example of which is depicted in Figure ~\ref{fig:skeleton}. The human skeleton we use follows the paradigm of \cite{sun2017compositional} expect that we define \emph{thorax} as root in all our experiments, in which 15 body parts are defined. \emph{Skeleton map}, like body part segmentation map, encodes part relationship in the body segments and imposes strong prior on human pose. However, training networks for full human body semantic segmentation needs labor-intensive dense and precise annotations, which is impractical for large human pose dataset \eg Human3.6M \cite{ionescu2014human3}. The simplicity of \emph{skeleton map} naturally addresses this issue. But what is a good architecture for \emph{skeleton map} generation?

\subsection{Deconvolution for Generating Skeleton Map}
\label{sec:deconv}
\begin{figure*}
\begin{center}
   \includegraphics[width=1.0\linewidth]{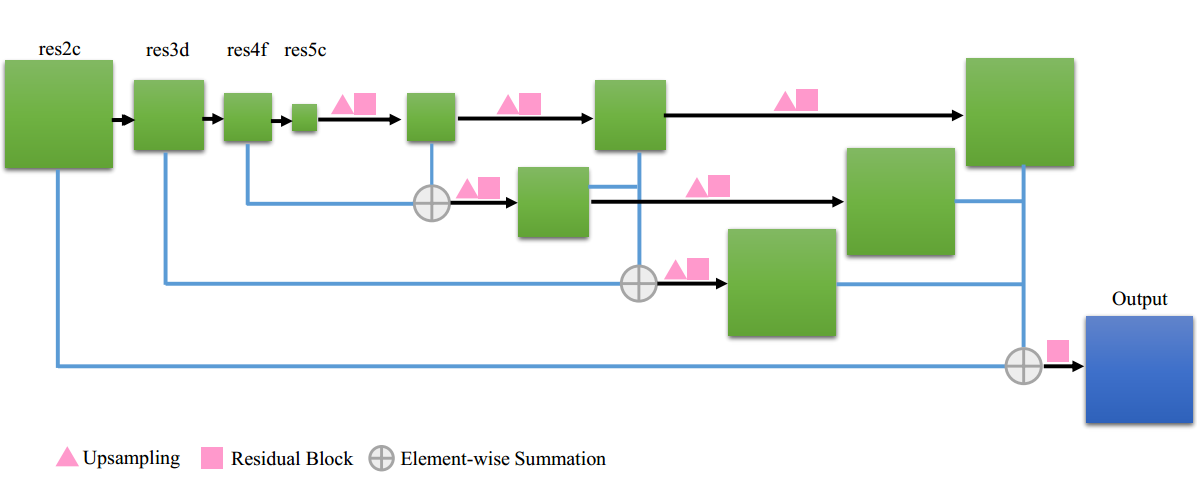}
\end{center}
   \caption{Deconvolutional Network Architecture. The blocks from input image up to \emph{res2c} are not drawn for simplicity. Intermediate supervision is not plotted. See Section \ref{sec:deconv} for details.}
\label{fig:diagram}

\end{figure*}

\textbf{Deconvolutional Network Design}
A simple choice is to employ the encoder-decoder architecture. In practice, we apply ResNet-50\cite{he2016deep} in a fully convolutional manner, producing pixel-level real values. We replace the fully connected layer after \emph{pool5} with deconvolutional layers. The network structure shown in Figure ~\ref{fig:diagram} starts with a 224$\times$224 image and extracts features along the downsampling process. Herein only \emph{res2c}, \emph{res3d}, \emph{res4f} and \emph{res5c} are sketched for brevity. The last fully connected layer is simply discarded. The devolutional module built upon \emph{pool5} gradually processes feature maps to the final output: a three-channel 56$\times$56 \emph{skeleton map}. It comprises of repeated blocks of upsampling layer (with initial weights set to bilinear upsampling) followed by residual module. To encourage the learning of local and global context, high-level feature maps are combined with low-level feature maps through skip connections, analogous to those used in deep networks\cite{long2015fully, newell2016stacked}. The output composes of three channels representing the \emph{skeleton map}. Rather than performing per-pixel classification into body part classes, we found that per-pixel regression results in better segmentation accuracy. We opt for sigmoid cross entropy loss in training. However, one common problem in training deep network is the notorious \emph{vanishing gradient}. To remedy this issue, each blob feeding into the pixel-wise summation layer branches off and connects to a residual block. Intermediate supervision is then applied on the output of each residual block, allowing for learning \emph{skeleton map} at multiple resolutions.

After the initial performance gain brought by common pratice \ie skip connection and intermediate supervision, investigations on prevalant \emph{conv-deconv} network architectures \eg \emph{RefineNet}\cite{lin2016refinenet} and \emph{Stacked Hourglass}\cite{newell2016stacked} did little to further improve the segmentation accuracy.

\textbf{Deal with trunction}
Truncation, that is, partial visibility of human body joints caused by image boundary, poses a noticeable challenge especially for in-the-wild 3D human pose estimation. In our case, this is supported by the fact that the deconvolutional network is uncertain about whether to plot the segments associated with cropped endpoint joint \ie \emph{wrist} or \emph{ankle}, due to lack of image evidence. A standard way to deal with this is to use multiple image crop, which is made possible by multiplying the provided rough person scale in the dataset with the rescaling factor \ie $c_i \in \left\lbrace 1.0, 1.25, 1.5 \right\rbrace$. The 2D joint ground truth is rescaled in the cropped window accordingly. Note that indoor dataset faces no truncation problem, and \emph{crop scale $c_i$} is always set to 1.0.

\textbf{Deal with stick width}
Several endeavours have been made to effectively excavate features for human pose estimation, most of which focus on \emph{multi-stream} learning\cite{xie2015holistically}. \emph{DeepSkeleton} differs in that it directly modifies the target output by changing stick width. Our design of \emph{multi-scale skeleton map} is motivated by the claim that each convolutional layer is responsible for skeleton pixel whose scale is less than the receptive field\cite{shen2016object}. In concrete terms, only convolutional layers with receptive field size larger than stick width can capture features of body parts. Hence, coarse segmentation(\emph{large stick width}) and fine segmentation(\emph{small stick width}) feature varying combinations of low-level and high-level features. Note that care has to be taken when deciding the body segment width of ground truth \emph{skeleton map}. We consider two practial concerns:(1)\emph{the parts should be small enough to localize semantic body parts.} (2)\emph{the parts should be large enough for convolution and deconvolution.}
The stick width \ie $l_i$ is empirically set to be in the range $[5,15]$.  

\begin{figure}
\begin{center}
   \includegraphics[width=0.19\linewidth]{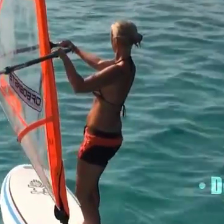}
   \includegraphics[width=0.19\linewidth]{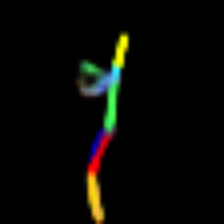}
   \includegraphics[width=0.19\linewidth]{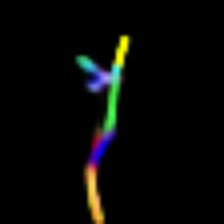}
   \includegraphics[width=0.19\linewidth]{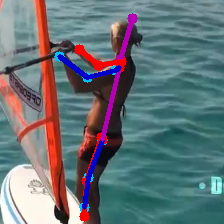}   
   \includegraphics[width=0.19\linewidth]{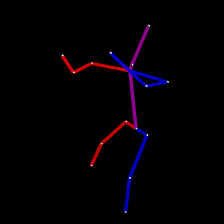}   
\end{center}
   \caption{A example of occlusion handling. From left to right: raw image, predicted \emph{foreground skeleton map}, predicted \emph{background skeleton map}, inferred 2D landmark, inferred 3D joint position. Note how the left/right body parts are distinguishably segmented in \emph{foreground/background skeleton map}.}
\label{fig:forebackground}

\end{figure}

\textbf{Deal with occlusion}
Severe occlusion hinders accurate human pose estimation. Answers to simple queries such as \emph{``Is the left upper leg in front of right upper leg?"} can actually provide important occlusion cues for regression. Motivated by \emph{inside/outside score map} \cite{li2016fully}, in this work, \emph{foreground skeleton map} $S_i^f$ displays body parts that are occluding others, while \emph{background skeleton map} $S_i^b$ displays those occluded by others. That said, \emph{skeleton map} explicitly models the occlusion relationship of two overlapping body parts. As far as we know, this straightforward formulation has never been done in the literature of human pose estimation. In more detail, recall that each 2D endpoint joint on the bone results in a ray oriented towards the camera optical center. Assume that 3D point $\left\lbrace \mathbf{X}_u,\mathbf{Y}_u,\mathbf{Z}_u\right\rbrace$ on bone $\mathbf{B}_u$ and $\left\lbrace \mathbf{X}_v,\mathbf{Y}_v,\mathbf{Z}_v \right\rbrace$ on bone $\mathbf{B}_v$ yield the same 2D projection $\left\lbrace \mathbf{x},\mathbf{y} \right\rbrace$. Denote the point with smaller depth(closer to camera in $Z$ direction) as $id$, and the other as $\widetilde{id}$.  \emph{Foreground skeleton map} assigns color of bone $\mathbf{B}_{id}$ to the pixel $\left\lbrace \mathbf{x},\mathbf{y} \right\rbrace$. In contrast, \emph{background skeleton map} assigns color of bone $\mathbf{B}_{\widetilde{id}}$, pretending bone $\mathbf{B}_{\widetilde{id}}$ is occluding bone $\mathbf{B}_{id}$. See Figure \ref{fig:forebackground} for an example.
This inherent \emph{occlusion-aware} property of \emph{skeleton map} is important for regression.

\subsection{Regression}
\label{sec:regress}
Plagued by the cluttered background of RGB image, a longstanding research direction in 3D human pose estimation has been exploiting better features from raw RGB input. We show that regression from \emph{skeleton map} alone is feasible. We employ state-of-the-art ResNet-50\cite{he2016deep} as the backbone network. Since \emph{skeleton maps} $(S_i^b, S_i^f)$ generated by deconvolutional network ${Deconv}_i$ are $56\times56$, they are rescaled to $224\times224$ first and concatenated together, which is then taken as input to ${Deconv}_i$ and processed along the downsampling path. The last fully connected layer is repurposed to output 3D position $X^i=\left\lbrace \mathbf{X}_k^i | k=1,2,..K \right\rbrace$ of all $K$ joints. Euclidean distance loss is applied for back propogation. Multiple 3D predictions $\mathcal{H}=\left\lbrace X^i | i=1,...n \right\rbrace $ are made by training independent regression networks for different \emph{skeleton maps} input. We want to emphasize that na\"{i}vely concatenating all the \emph{skeleton maps} $\mathcal{S}$ as input (\emph{early fusion}) or learning parallel streams for multiple input and fusing the feature responses subsequently (\emph{late fusion}) will not help. As an alternative option, one might consider to concatenate \emph{skeleton map} along with raw RGB image, which however, does not boost the performance in our observation. Therefore, we stick with the original design \ie learning 3D solely from intermediate \emph{skeleton map} feature representation.

\subsection{Matching}
\label{sec:matching}
Now we have multiple 3D pose hypotheses $\mathcal{H}$, the problem boils down to select the optimal hypothesis $X^{\ast} \in \mathcal{H}$ as final 3D output. The simpliest way is to choose the candidate whose projection best matches the 2D pose detection results. Writing $\mathit{Proj}$ as the camera projection matrix and $x$ as 2D pose detection, we seek to find the optimal 3D pose $X^{\ast} \in \mathcal{H}$ via minimizing the reprojection error:
\begin{equation}
X^{\ast}=\underset{X^i}{\operatorname{arg\,min}}||Proj(X^i)-x||^2
\end{equation}

We use the pre-trained state-of-the-art 2D detector \emph{Stacked Hourglass}\cite{newell2016stacked} for generating $x$. No finetuning is employed. We remark that our 3D hypotheses are completely independent of 2D pose detection, rather they are learnt from multi-level discriminative \emph{skeleton maps}. 

\section{Discussions}
In principle, any expressive intermediate representation can be used for regression. Heatmap, for example, has been explorered in \cite{wu2016single} to bridge the gap between real and synthetic dataset. Carreira \etal \cite{carreira2016human} stack heatmap with RGB image for coarse-to-fine regression. Yet, an obvious drawback of heatmap is different body joints are encoded in \emph{discrete gaussian peaks}, thus the dependence between joints is not well exploited. \emph{Skeleton map} overcomes this problem by explicitly connecting adjacent joints of each bone by a stick. The colored semantic body part offers a strong cue for regression learning. For generating \emph{skeleton map}, perhaps the easiest way is to firstly detect 2D joints and then draw lines between neighbouring joints. However, this introduces two disadvantages: \emph{1. Occlusion relationship information is completely discarded. 2. The inaccurate 2D detection has impact on the following regression.} Our initial exploration shows that this has no apparent benefits.
\section{Experiments}
Our approach is evaluated on the largest human pose benchmarks MPII and Human3.6M. 
\emph{MPII} \cite{andriluka20142d} is a 2D real-world human pose dataset. It contains around $25k$ natural images collected from YouTube with a variety of poses and complicated image appearances. Cluttered background, multiple people, severe truncation and occlusion make it the most challenging 2D human pose dataset. For 2D pose evaluation in MPII, we use PCKh\cite{andriluka20142d} metric which measures the percentage of joints with distance to ground truth below a certain threshold.

\emph{Human3.6M}\cite{ionescu2014human3} is a large-scale 3D human pose dataset consisting of 3.6M video frames captured in controlled laboratory environment. 5 male and 6 female actors performing 17 daily activities are captured by motion capture system in 4 different camera views. The image appearance in clothes and background is limited compared to \emph{MPII}. Following standard practice in \cite{zhou2017towards, li20143d, zhou2016deep, zhou2016sparseness}, five subjects(S1, S5, S6, S7, S8) are used in training. Every $64^{th}$ frame of the two subjects(S9, S11) is used in testing. MPJPE(mean per joint position error)\cite{ionescu2014human3} is used as evaluation metric after aligning 3D poses to root joint. We represent 3D pose in local camera coordinate system following the methodology of Zhou \etal \cite{zhou2017towards}.
\subsection{Implementation Detail}
For training the network we use Caffe\cite{jia2014caffe} with 15 GPUs. Deconvolutional network training starts with base learning rate 0.00001 and mini-batch size 12. For training regression network, we set base learning rate to 0.01 and mini-batch size to 32. Learning rate is dropped by a factor of 10 after error plateu on the validation set. The network is trained until convergence. For optimization, stochastic gradient descent is adopted. Weight decay is 0.0002, and momentum is 0.9. No data augmentation or data fusion is used. 

\subsection{Baseline Settings}
To validate the effectiveness of \emph{skeleton map} and \emph{multiple hypothesis}, we test two baselines:
\begin{itemize}
\item[$\bullet$] \emph{Direct RGB} 

It performs regression directly from raw RGB input.
\item[$\bullet$] \emph{Ours w/o Mul-Hyp} 

It performs regression from only one \emph{skeleton map}.
\end{itemize}

Unless otherwise specified, we set $c_i=1.0,l_i=10$ for all the experiments of one hypothesis(\emph{skeleton map}) both on Human3.6M and MPII.

Our final system is denoted as \emph{Ours w Mul-Hyp} (equivalent to \emph{DeepSkeleton}).
\subsection{Experiments on 3D dataset Human3.6M}
\label{sec:h36mexp}
\begin{figure*}
\begin{center}
   \includegraphics[width=0.075\linewidth]{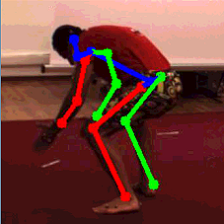}
   \includegraphics[width=0.075\linewidth]{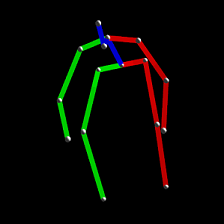}
   \includegraphics[width=0.075\linewidth]{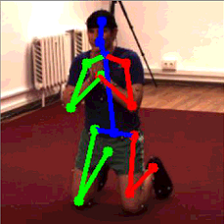}   
   \includegraphics[width=0.075\linewidth]{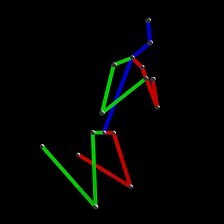}
   \includegraphics[width=0.075\linewidth]{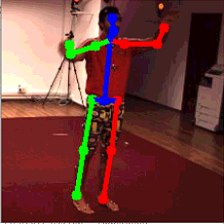}
   \includegraphics[width=0.075\linewidth]{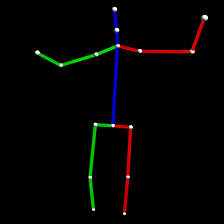}   
   \includegraphics[width=0.075\linewidth]{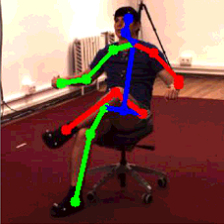}
   \includegraphics[width=0.075\linewidth]{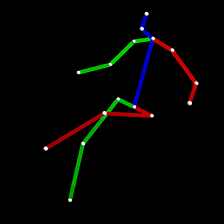}
   \includegraphics[width=0.075\linewidth]{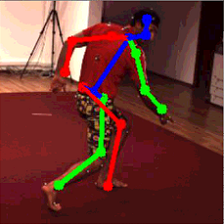}   
   \includegraphics[width=0.075\linewidth]{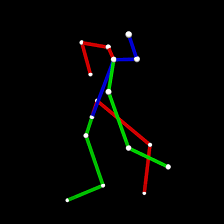}   
   \includegraphics[width=0.075\linewidth]{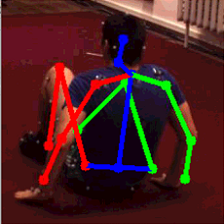}
   \includegraphics[width=0.075\linewidth]{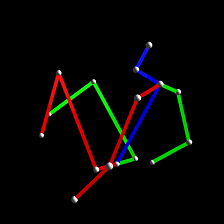}
   
   \includegraphics[width=0.075\linewidth]{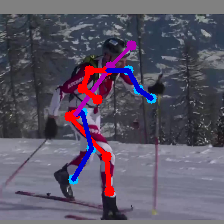}
   \includegraphics[width=0.075\linewidth]{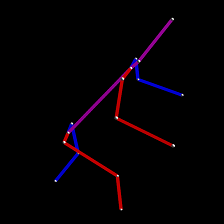}
   \includegraphics[width=0.075\linewidth]{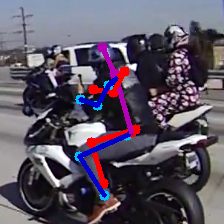}   
   \includegraphics[width=0.075\linewidth]{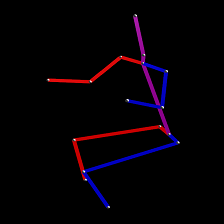}
   \includegraphics[width=0.075\linewidth]{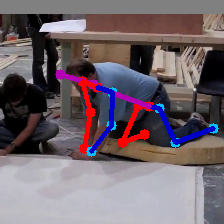}
   \includegraphics[width=0.075\linewidth]{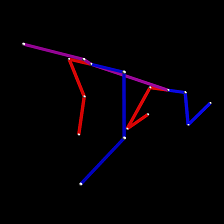}   
   \includegraphics[width=0.075\linewidth]{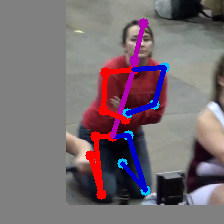}
   \includegraphics[width=0.075\linewidth]{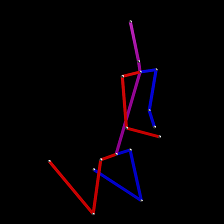}
   \includegraphics[width=0.075\linewidth]{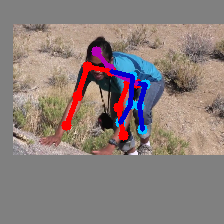}   
   \includegraphics[width=0.075\linewidth]{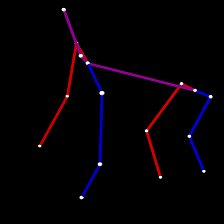}
   \includegraphics[width=0.075\linewidth]{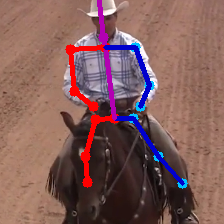}
   \includegraphics[width=0.075\linewidth]{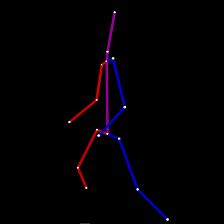}
   
   \includegraphics[width=0.075\linewidth]{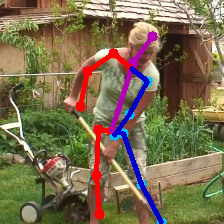}
   \includegraphics[width=0.075\linewidth]{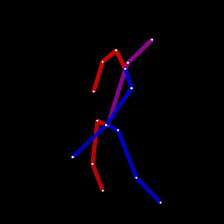}
   \includegraphics[width=0.075\linewidth]{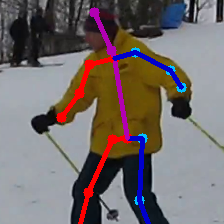}   
   \includegraphics[width=0.075\linewidth]{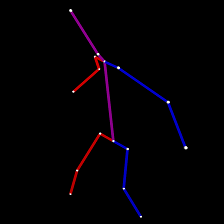}
   \includegraphics[width=0.075\linewidth]{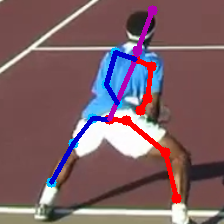}
   \includegraphics[width=0.075\linewidth]{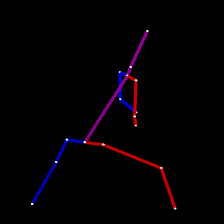}   
   \includegraphics[width=0.075\linewidth]{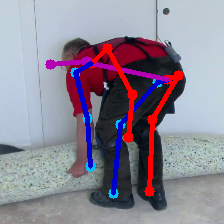}
   \includegraphics[width=0.075\linewidth]{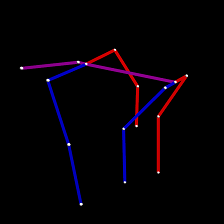}
   \includegraphics[width=0.075\linewidth]{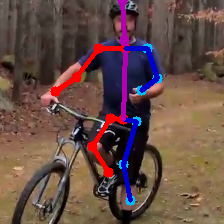}   
   \includegraphics[width=0.075\linewidth]{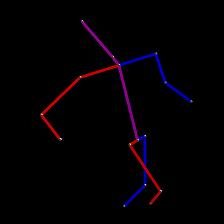}
   \includegraphics[width=0.075\linewidth]{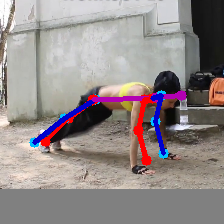}
   \includegraphics[width=0.075\linewidth]{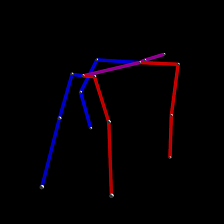}
   
   \includegraphics[width=0.075\linewidth]{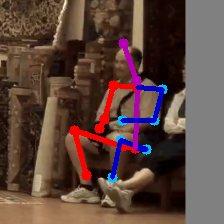}
   \includegraphics[width=0.075\linewidth]{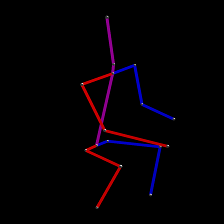}                              
   \includegraphics[width=0.075\linewidth]{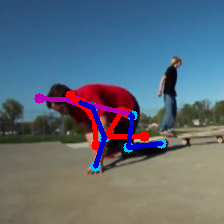}      
   \includegraphics[width=0.075\linewidth]{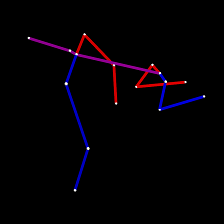}  
   \includegraphics[width=0.075\linewidth]{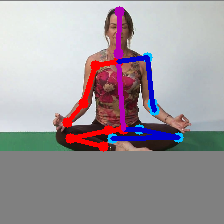}
   \includegraphics[width=0.075\linewidth]{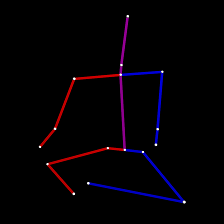}      
   \includegraphics[width=0.075\linewidth]{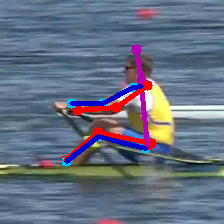}   
   \includegraphics[width=0.075\linewidth]{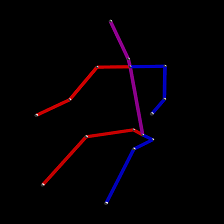}   
   \includegraphics[width=0.075\linewidth]{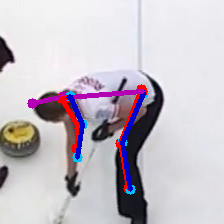}      
   \includegraphics[width=0.075\linewidth]{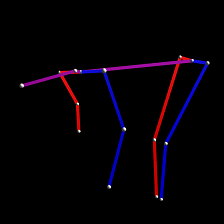}   
   \includegraphics[width=0.075\linewidth]{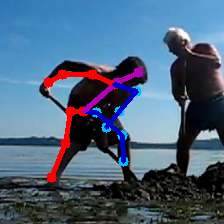}   
   \includegraphics[width=0.075\linewidth]{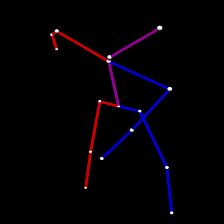}   
\end{center}
   \caption{Qualitative results on Human3.6M(First row) and MPII(Second to fourth row). 3D poses are illustrated from a novel viewpoint. Note that 3D pose results for natural images(MPII) are quite plausible. Different colors are used to differentiate MPII from Human3.6M.}
\label{fig:qualitative}

\end{figure*}

\textbf{Comparison with state-of-the-art}
For fair comparison, we compare with state-of-the-art methods without mixed 2D and 3D data for training in Table \ref{table:compstateofthearth36m}. Note that \emph{Compositional Pose Regression} \cite{sun2017compositional} provides results with and without 2D data. We therefore denote $\emph{CompBone}^* $ as \emph{Compositional Pose Regression} without extra 2D training data and report the results from the original paper. Table \ref{table:compstateofthearth36m} shows that our final system \emph{ours w Mul-Hyp} outperforms the main competitor Tome \etal \cite{tome2017lifting}. Notably, it surpasses competeting methods in actions \emph{Sit}, \emph{SitDown}, \emph{Photo} by a large margin. The improvement comes from our novel \emph{skeleton map} representation and the expressiveness of multiple hypotheses. Visualized 3D poses are displayed in Figure \ref{fig:qualitative}.

\begin{table*}
\scriptsize
\begin{center}
\begin{tabular}{lcccccccc}

\toprule

\textbf{Method} & \textbf{Direction} & \textbf{Discuss} & \textbf{Eat} & \textbf{Greet} & \textbf{Phone} & \textbf{Pose} & \textbf{Purchase} & \textbf{Sit} \\
\hline\
Tekin\cite{tekin2016direct} & 102.4 & 147.7 & 88.8 & 125.3 & 118.0 & 112.4 & 129.2 & 138.9\\
Chen\cite{chen20163d} & 89.9 & 97.6 & 90.0 & 107.9 & 107.3 & 93.6 & 136.1 & 133.1\\
Zhou\cite{zhou2016sparseness} & 87.4 & 109.3 & 87.1 & 103.2 & 116.2 & 106.9 & 99.8 & 124.5\\
Xingyi\cite{zhou2016deep} & 91.8 & 102.4 & 97.0 & 98.8 & 113.4 & 90.0 & 93.8 & 132.2\\
${CompBone}^{*}$\cite{sun2017compositional} & 90.2 & 95.5 & 82.3 & 85.0 & 87.1 & 87.9 & 93.4 & 100.3 \\
Tome\cite{tome2017lifting} & \textbf{65.0} & \textbf{73.5} & \textbf{76.8} & 86.4 & \textbf{86.3} & \textbf{68.9} & \textbf{74.8} & 110.2\\
Moreno-Noguer\cite{moreno20163d} & 69.5 & 80.2 & 78.2 & 87.0 & 100.8 & 76.0 & 69.7 & 104.7\\
Ours w Mul-Hyp & 75.6 & 75.0 & 94.9 & \textbf{82.4} & 107.7 & 91.2 & 86.6 & \textbf{73.3}\\
\hline
\toprule
\textbf{Method} & \textbf{SitDown} & \textbf{Smoke} & \textbf{Photo} & \textbf{Wait} & \textbf{Walk} & \textbf{WalkDog} & \textbf{WalkTogether} & \textbf{Avg}  \\
\hline\
Tekin\cite{tekin2016direct} & 224.9 & 118.4 & 182.7 & 138.8 & 55.1 & 126.3 & 65.8 & 125.0\\
Chen\cite{chen20163d} & 240.1 & 106.7 & 139.2 & 106.2 & 87.0 & 114.1 & 90.6 & 114.2\\
Zhou\cite{zhou2016sparseness} & 199.2 & 107.4 & 143.3 & 118.1 & 79.4 & 114.2 & 97.7 & 113.0\\
Xingyi\cite{zhou2016deep} & 159.0 & 106.9 & 125.2 & 94.4 & 79.0 & 126.0 & 99.0 & 107.3\\
${CompBone}^{*}$\cite{sun2017compositional} & 135.4 & 91.4 & 94.5 & 87.3 & 78.0 & 90.4 & 86.5 & 92.4 \\
Tome\cite{tome2017lifting} & 173.9 & 85.0 & 110.7 & \textbf{85.8} & \textbf{71.4} & 86.3 & \textbf{73.1} & 88.4\\
Moreno-Noguer\cite{moreno20163d} & 113.9 & 89.7 & 102.7 & 98.5 & 79.2 & 82.4 & 77.2 & 87.3\\
Ours w Mul-Hyp & \textbf{80.5} & \textbf{83.9} & \textbf{81.5} & 97.1 & 99.4 & \textbf{78.9} & 87.0 & \textbf{86.5}\\
\bottomrule
\end{tabular}
\end{center}
\caption{Comparison with state-of-the-art on Human3.6M. No mixed 2D and 3D data training is used in all the methods. MPJPE(mean per joint position error) is used as evaluation metric.}
\label{table:compstateofthearth36m}
\end{table*}

\textbf{Comparison with Regression from RGB}
Table \ref{table:directrgb} shows that \emph{Ours w/o Mul-Hyp} significantly improves baseline \emph{Direct RGB} by 12.3\emph{mm}(relative 10.8\%), demonstrating the strength of \emph{skeleton map}.

\begin{table}
\scriptsize
\begin{center}
\begin{tabular}{ll}
\toprule
\textbf{Method} & \textbf{Avg MPJPE} \\
\hline\
Direct RGB & 114.2\\
Ours w/o Mul-Hyp & $101.9_{\downarrow 12.3}$\\
Ours w Mul-Hyp & ${\textbf{86.5}}_{\downarrow 27.7}$\\

\bottomrule
\end{tabular}
\end{center}
\caption{Comparison of ours with regression from raw RGB on Human3.6M. Mul-Hyp=multiple hypotheses. MPJPE  metric is used.}
\label{table:directrgb}
\end{table}

\emph{Does skeleton map brings better 2D estimation, or better depth estimation?} In order to answer this question, we evaluate the average joint error given ground truth depth and ground truth 2D respectively in the following. Without loss of generality, we restrict ourselves to generate one hypothesis.

\textbf{Impact of Skeleton Map on 2D Estimation} We make use of ground truth depth and predicted 2D to recover 3D pose in the camera coordinate system. Table \ref{table:module2D} reports the result of average joint error using different input sources for regression network. One can see that 25.9\emph{mm}(relative 27.9\%) error reduction is obtained after feeding predicted \emph{skeleton map} to regression network. Further 17.4\emph{mm} decrease is achieved by using ground truth \emph{skeleton map} for regression. This can be interpreted as \emph{skeleton map} simplifies the 2D learning procedure and prevents overfitting. Strong shape prior serves as important regularization cue for learning 2D location.

\begin{table}
\scriptsize
\begin{center}
\begin{tabular}{ll}
\toprule
\textbf{Method} & \textbf{Avg MPJPE} \\
\hline\
Direct RGB w GT Depth, Pred 2D & 92.8\\
Pred Ske w GT Depth, Pred 2D & $66.9_{\downarrow 25.9}$\\
GT Ske w GT Depth, Pred 2D & ${\textbf{49.5}}_{\downarrow 43.3}$\\
\bottomrule
\end{tabular}
\end{center}
\caption{Performance given ground truth depth of different regression input on Human3.6M. Pred(GT) Ske=Use predicted(ground truth) \emph{skeleton map} for regression. Direct RGB=Use RGB for regression. Pred 2D=Use predicted 2D. GT Depth=Use ground truth depth. MPJPE metric is used.} 
\label{table:module2D}
\end{table}

\textbf{Impact of Skeleton Map on Depth Estimation}
To gain insight into the importance of \emph{skeleton map} for depth estimation, we use ground truth 2D and predicted depth to acquire 3D joints. We see in Table \ref{table:moduleDepth} that depth regression from predicted \emph{skeleton map} shows evident superiority over RGB image, yielding 21.5\emph{mm}(relative 22.7\%) error reduction. This indicates that \emph{skeleton map} is more favorable for depth prediction.

\begin{figure}
\begin{center}
   \includegraphics[width=1.0\linewidth]{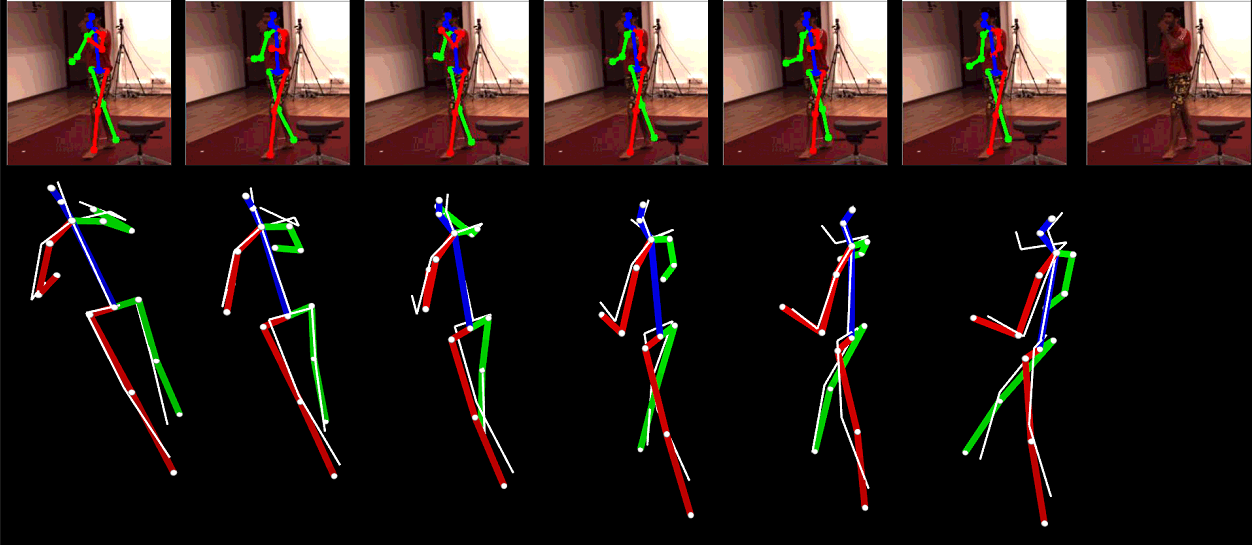}
\end{center}
   \caption{Visualization of multiple hypotheses on Human3.6M. Bottom shows the predicted hypotheses and ground truth hypotheses(white) from a novel viewpoint. Top shows the projection of 3D hypotheses and raw image. The third hypothesis is the final output.}
\label{fig:multihyp}

\end{figure}

\textbf{Impact of Multiple Hypotheses} Next we elaborate on the effect of using multiple hypotheses. Here we use $k=11$ hypotheses.\footnote{$c_i=1.0, l_i \in \lbrace 5, 6, 7, 8, 9, 10, 11, 12, 13, 14, 15 \rbrace$} We first assume that the ground truth \emph{skeleton map} is provided. In Table \ref{table:multiGTSke}, multiple hypotheses slightly improves the accuracy, but to a lower extent than expected. This implies that ground truth \emph{skeleton map} is sufficiently powerful to reduce ambiguity. We then move to a realistic scenario where ground truth \emph{skeleton map} is unavailable. Quite surprisingly, using multiple hypotheses reduces the average MPJPE from 101.9 \emph{mm} to 86.5\emph{mm} in Table \ref{table:directrgb}, which largely narrows down the performance gap between ground truth and predicted \emph{skeleton map}. Generated multiple hypotheses are illustrated in Fig \ref{fig:multihyp}. The third hypothesis is chosen as final output based on simple matching. One could argue that similar performance might be accomplished by ensembling multiple runs of the same regression network ${Regression}_i$. To examine this, we take the regression outputs of $k=11$ different runs from single \emph{skeleton map}, denoted as \emph{Ensemble}. The result in Table \ref{table:multiPredSke} suggests that our multi-level \emph{skeleton map} provides more information than single \emph{skeleton map}. A natural problem arises: \emph{What is the performance upper bound of multiple hypotheses?}  To investigate this problem, we select the optimal 3D hypothesis with minimum 3D error to ground truth 3D pose, producing an error of 68.3$mm$. This is promising as we are able to excel most state-of-the-art works \emph{without offline 3D pose library}. However, how to select the optimal 3D hypothesis remains unclear.
\begin{table}
\scriptsize
\begin{center}
\begin{tabular}{ll}
\toprule
\textbf{Method} & \textbf{Avg MPJPE} \\
\hline\
Direct RGB w GT 2D, Pred Depth & 94.6\\
Pred Ske w GT 2D, Pred Depth & $73.1_{\downarrow 21.5}$\\
GT Ske w GT 2D, Pred Depth & $\textbf{58.7}_{\downarrow 35.9}$\\
\hline
\end{tabular}
\end{center}
\caption{Performance given ground truth 2D of different regression input on Human3.6M. Pred(GT) Ske=Use predicted(ground truth) \emph{skeleton map} for regression. Direct RGB=Use RGB for regression. GT 2D=Use ground truth 2D. Pred Depth=Use predicted depth. MPJPE metric is used.} 
\label{table:moduleDepth}
\end{table}

\begin{table}
\scriptsize
\begin{center}
\begin{tabular}{ll}
\toprule
\textbf{Method} & \textbf{Avg MPJPE} \\
\hline\
GT Ske w/o Mul-Hyp & 76.2\\
GT Ske w Mul-Hyp & $\textbf{69.7}_{\downarrow 6.5}$\\
\bottomrule
\end{tabular}
\end{center}
\caption{Performance gain from multiple hypotheses given ground truth skeleton map on Human3.6M. MPJPE metric is used.} 
\label{table:multiGTSke}
\end{table}

\begin{table*}
\scriptsize
\begin{center}
\begin{tabular}{lllllllll}

\toprule

\textbf{Method} & \textbf{Direction} & \textbf{Discuss} & \textbf{Eat} & \textbf{Greet} & \textbf{Phone} & \textbf{Pose} & \textbf{Purchase} & \textbf{Sit} \\
\hline\
Ours w/o Mul-Hyp & 94.5 & 89.1 & 103.8 & 101.7 & 131.4 & 97.0 & 107.7 & 84.4\\
Ensemble & $89.7_{\downarrow 4.8}$ & $85.2_{\downarrow 3.9}$ & $100.3_{\downarrow 3.5}$ & $97.7_{\downarrow 4.0}$ & $126.5_{\downarrow 4.9}$ & $94.0_{\downarrow 3.0}$ & $102.2_{\downarrow 5.5}$ & $80.7_{\downarrow 3.7}$\\
Ours w Mul-Hyp & $\textbf{75.6}_{\downarrow 18.9}$ & $\textbf{75.0}_{\downarrow 14.1}$ & $\textbf{94.9}_{\downarrow 8.9}$ & $\textbf{82.4}_{\downarrow 19.3}$ & $\textbf{107.7}_{\downarrow 23.7}$ & $\textbf{91.2}_{\downarrow 5.8}$ & $\textbf{86.6}_{\downarrow 21.1}$ & $\textbf{73.3}_{\downarrow 11.1}$\\
\hline
\toprule
\textbf{Method} & \textbf{SitDown} & \textbf{Smoke} & \textbf{Photo} & \textbf{Wait} & \textbf{Walk} & \textbf{WalkDog} & \textbf{WalkTogether} & \textbf{Avg}  \\
\hline\
Ours w/o Mul-Hyp & 97.7 & 93.8 & 98.0 & 111.6 & 110.9 & 97.6 & 111.0 & 101.9\\
Ensemble & $95.0_{\downarrow 2.7}$ & $90.9_{\downarrow 2.9}$ & $94.0_{\downarrow 4.0}$ & $104.6_{\downarrow 7.0}$ & $106.8_{\downarrow 4.1}$ & $92.6_{\downarrow 5.0}$ & $106.6_{\downarrow 4.4}$ & $97.7_{\downarrow 4.2}$\\
Ours w Mul-Hyp & $\textbf{80.5}_{\downarrow 17.2}$ & $\textbf{83.9}_{\downarrow 9.9}$ & $\textbf{81.5}_{\downarrow 16.5}$ & $\textbf{97.1}_{\downarrow 14.5}$ & $\textbf{99.4}_{\downarrow 11.5}$ & $\textbf{78.9}_{\downarrow 18.7}$ & $\textbf{87.0}_{\downarrow 24.0}$ & $\textbf{86.5}_{\downarrow 15.4}$ \\
\bottomrule
\end{tabular}
\end{center}
\caption{Performance gain from multiple hypotheses given predicted \emph{skeleton map} on Human3.6M. MPJPE metric is used. }
\label{table:multiPredSke}
\end{table*}

\subsection{Experiments on 2D dataset MPII}
We present 2D and 3D pose estimation for in-the-wild dataset MPII. We use MPII validation set \cite{tompson2014joint} including 2958 images for ablation study.

\textbf{Pseudo 3D Ground Truth}
MPII only provides 2D annotation, but training our network requires 3D pose ground truth. We use state-of-the-art 3D reconstruction approach \cite{zhou2017sparse} to initialize 3D pose from 2D landmark.
Note that most of the reconstructed poses are already reasonable despite occasional incorrect inference. We then introduce human assistance, where a human expert is presented with the initialized 3D pose along with input image and asked to manually adjust wrong limb orientation. We stress that the goal of semi-automatic annotation is to resolve the depth ambiguity as far as possible by aligning 3D pose with image observation. Since accurate 3D MoCap pose is impratical for natural images, we call this \emph{pseudo 3D ground truth}.

\begin{table}
\scriptsize
\begin{center}
\begin{tabular}{lcccccccc}
\toprule
\textbf{Method} & \textbf{All}\\
\hline\
Wei\cite{wei2016convolutional} & 88.5\\
Newell\cite{newell2016stacked} & 90.9\\
Chu\cite{chu2017multi} & \textbf{91.5}\\

\toprule
Carreira(IEF)\cite{carreira2016human} & 81.3\\

\toprule
Sun(CompBone)\cite{sun2017compositional} & \textbf{86.4}\\
${CompBone}^{**}$ & 79.6\\
Rogez(LCR-Net)\cite{rogez2017lcr} & 74.2\\
Ours w Mul-Hyp & 73.1 \\
\bottomrule
\end{tabular}
\end{center}
\caption{Comparison with state-of-the-art on MPII test set. PCKh@0.5 is used as evaluation metric. All denotes PCKh@0.5 of all joints. Top section: 2D detection based. Middle section: 2D regression based. Bottom section: 3D regression based.} 
\label{table:compstateoftheartpckh}
\end{table}

\textbf{Comparison with state-of-the-art} 
Previous approaches generally fall into three families: 2D detection based, 2D regression based and 3D regression based. Our method belongs to 3D regression based. In this family, our closest competitors are \cite{rogez2017lcr, sun2017compositional}. The comparison is not completely fair as they both use additional 3D data. Sun \etal \cite{sun2017compositional} integrate a two-stage state-of-the-art 2D regression based method IEF\cite{carreira2016human} into their network. For the completeness of this work, we also report Stage 0 result provided in their paper, denoted as $\emph{CompBone}^{**} $. This amounts to direct regression without ad-hoc stage. We observe in Table \ref{table:compstateoftheartpckh} \emph{ours w Mul-Hyp} is on par with state-of-the-art 3D regression methods without 3D data or post processing. Qualitative results are shown in Figure \ref{fig:qualitative}.

\textbf{Comparison with Regression from RGB} Table \ref{table:compwithbaselinepckh} compares \emph{Ours w/o Mul-Hyp} with \emph{Direct RGB}. We observe that each joint gains tremendous improvement. For instance, elbow PCKh@0.5 is improved by 12.0\%(relative 20.7\%) and ankle PCKh@0.5 is improved by 5.7\%(relative 17.2\%). This again demonstrates the remarkable merit of \emph{skeleton map}.

\begin{table}
\scriptsize
\begin{center}
\begin{tabular}{lllll}

\toprule
\textbf{Method} & \textbf{Head} & \textbf{Sho.} & \textbf{Elb.} & \textbf{Wri.}\\
\hline\
Direct RGB & 79.1 & 75.1 & 58.0 & 46.9 \\
Ours w/o Mul-Hyp & $\textbf{90.6}_{\uparrow 11.5}$ & $\textbf{83.4}_{\uparrow 8.3}$ & $\textbf{70.0}_{\uparrow 12.0}$ & $\textbf{54.5}_{\uparrow 7.6}$ \\
\toprule
\textbf{Method} & \textbf{Hip} & \textbf{Knee} & \textbf{Ank.} & \textbf{Mean} \\
\hline\
Direct RGB & 64.5 & 49.0 & 33.1 & 61.8\\
Ours w/o Mul-Hyp & $\textbf{74.2}_{\uparrow 9.7}$ & $\textbf{59.4}_{\uparrow 10.4}$ & $\textbf{38.8}_{\uparrow 5.7}$ & $\textbf{71.2}_{\uparrow 9.4}$\\

\bottomrule
\end{tabular}
\end{center}
\caption{Comparison to direct regression from RGB on MPII validation set. Regression from only one \emph{skeleton map} increases mean PCKh@0.5 by 9.4$\%$.} 
\label{table:compwithbaselinepckh}
\end{table}

\textbf{Impact of Multiple Hypotheses} Table \ref{table:mpiimultihyp} shows the effect of multi-scale and multi-crop \emph{skeleton map}. We observe the same conclusion as in Table \ref{table:multiPredSke}. Using multi-scale \emph{skeleton map} results in 9.4\%(relative 24.2\%) improvement of ankle PCKh@0.5. Multi-crop \emph{skeleton map} yields extra 7.0\% improvement. 
It is noteworthy that ensemble of the same regression network falls behind our final system, indicating multi-level \emph{skeleton map} is able to capture diverse semantic features from input image. 

\begin{table}
\scriptsize
\begin{center}
\begin{tabular}{lllll}
\toprule
\textbf{Method} & \textbf{Head} & \textbf{Sho.} & \textbf{Elb.} & \textbf{Wri.} \\
\hline\
Base & 90.6 & 83.4 & 70.0 & 54.5 \\
Ensemble & $89.4_{\downarrow 1.2}$ & $83.7_{\uparrow 0.3}$ & $66.9_{\downarrow 3.1}$ & $54.8_{\uparrow 0.3}$ \\
Base+Mul-S & $90.1_{\downarrow 0.5}$ & $85.6_{\uparrow 2.2}$ & $\textbf{70.7}_{\uparrow 0.7}$ & $57.2_{\uparrow 2.7}$ \\
Base+Mul-S+Mul-C & $\textbf{90.5}_{\uparrow 0.2}$ & $\textbf{86.0}_{\uparrow 2.6}$ & ${70.5}_{\uparrow 0.5}$ & $\textbf{58.0}_{\uparrow 3.5}$ \\

\toprule
\textbf{Method} & \textbf{Hip} & \textbf{Knee} & \textbf{Ank.} & \textbf{Mean} \\
\hline\
Base & 74.2 & 59.4 & 38.8 & 71.2\\
Ensemble & $73.5_{\downarrow 0.7}$ & $63.3_{\uparrow 3.9}$ & $48.9_{\uparrow 10.1}$ & $71.9_{\uparrow 0.7}$\\
Base+Mul-S & $\textbf{75.6}_{\uparrow 1.4}$ & $65.3_{\uparrow 5.9}$ & $48.2_{\uparrow 9.4}$ & $73.7_{\uparrow 2.5}$\\
Base+Mul-S+Mul-C & ${75.1}_{\uparrow 0.9}$ & $\textbf{68.4}_{\uparrow 9.0}$ & $\textbf{55.2}_{\uparrow 16.4}$ & $\textbf{74.8}_{\uparrow 3.6}$\\
\bottomrule
\end{tabular}
\end{center}
\caption{Comparison to ours with single hypothesis on MPII validation set. Base: \emph{Ours w/o Mul-Hyp}. Mul-S: Vary stick width of \emph{skeleton map} $l_i$ in $\lbrace 5, 6, 7, 8, 9, 10\rbrace $. Mul-C: Vary crop size of raw image $c_i$ in $\lbrace 1.0, 1.25, 1.5\rbrace $. Ensemble: 18 different runs of regression from one \emph{skeleton map}. PCKh@0.5 metric is used.} 
\label{table:mpiimultihyp}
\end{table}

\textbf{Performance Upperbound} One remaining question is \emph{what is the limit of skeleton map applied  in natural unconstrained scenario?} To assess the upper bound, we perform regression from one single ground truth \emph{skeleton map} on MPII. We see in Table \ref{table:compwithoursonepckh} regression from single ground truth \emph{skeleton map} achieves 94.5\% overall PCKh@0.5. This validates the effectiveness of \emph{skeleton map} representation.

\begin{table}
\scriptsize
\begin{center}
\begin{tabular}{lllll}
\toprule
\textbf{Method} & \textbf{Head} & \textbf{Sho.} & \textbf{Elb.} & \textbf{Wri.}\\
\hline\
Ours w/o Mul-Hyp & 90.6 & 83.4 & 70.0 & 54.5\\
GT Ske w/o Mul-Hyp & $\textbf{96.9}_{\uparrow 6.3}$ & $\textbf{99.4}_{\uparrow 16.0}$ & $\textbf{97.2}_{\uparrow 27.2}$ & $\textbf{93.4}_{\uparrow 38.9}$\\

\toprule

\textbf{Method} & \textbf{Hip} & \textbf{Knee} & \textbf{Ank.} & \textbf{Mean}  \\
\hline\
Ours w/o Mul-Hyp & 74.2 & 59.4 & 38.8 & 71.2\\
GT Ske w/o Mul-Hyp & $\textbf{97.7}_{\uparrow 23.5}$ & $\textbf{94.0}_{\uparrow 34.6}$ & $\textbf{70.1}_{\uparrow 31.3}$ & $\textbf{94.5}_{\uparrow 23.3}$\\

\bottomrule
\end{tabular}
\end{center}
\caption{Comparison of regression from one predicted \emph{skeleton map}(\emph{Ours w/o Mul-Hyp}) and regression from one ground truth \emph{skeleton map}(\emph{GT Ske w/o Mul-Hyp}) on MPII validation set. PCKh@0.5 metric is used.} 
\label{table:compwithoursonepckh}
\end{table}

\section{Conclusion}
We have sucessfully shown how to push the limit of 3D human pose estimation using \emph{skeleton map} without fusing different data sources. \emph{Skeleton map} is an impressive abstraction of input, which when combined with multiple hypotheses generation is able to achieve compelling results on both indoor and in-the-wild dataset. We also carry out exhaustive experimental evaluation to understand the performance upper bound of our novel intermediate representation. We expect to further narrow down the performance gap between ground truth and predicted \emph{skeleton map} by better segmentation network. We hope the idea of combining semantic segmentation and pose estimation inspire a new research direction in 3D human pose estimation.

{\small
\bibliographystyle{ieee}
\bibliography{egbib}

\begin{thebibliography}{10}\itemsep=-1pt

\bibitem{akhter2015pose}
I.~Akhter and M.~J. Black.
\newblock Pose-conditioned joint angle limits for 3d human pose reconstruction.
\newblock In {\em Proceedings of the IEEE Conference on Computer Vision and
  Pattern Recognition}, pages 1446--1455, 2015.

\bibitem{alahari2013pose}
K.~Alahari, G.~Seguin, J.~Sivic, and I.~Laptev.
\newblock Pose estimation and segmentation of people in 3d movies.
\newblock In {\em Proceedings of the IEEE International Conference on Computer
  Vision}, pages 2112--2119, 2013.

\bibitem{andriluka20142d}
M.~Andriluka, L.~Pishchulin, P.~Gehler, and B.~Schiele.
\newblock 2d human pose estimation: New benchmark and state of the art
  analysis.
\newblock In {\em Proceedings of the IEEE Conference on computer Vision and
  Pattern Recognition}, pages 3686--3693, 2014.

\bibitem{bogo2016keep}
F.~Bogo, A.~Kanazawa, C.~Lassner, P.~Gehler, J.~Romero, and M.~J. Black.
\newblock Keep it smpl: Automatic estimation of 3d human pose and shape from a
  single image.
\newblock In {\em European Conference on Computer Vision}, pages 561--578.
  Springer, 2016.

\bibitem{bulat2016human}
A.~Bulat and G.~Tzimiropoulos.
\newblock Human pose estimation via convolutional part heatmap regression.
\newblock In {\em European Conference on Computer Vision}, pages 717--732.
  Springer, 2016.

\bibitem{carreira2016human}
J.~Carreira, P.~Agrawal, K.~Fragkiadaki, and J.~Malik.
\newblock Human pose estimation with iterative error feedback.
\newblock In {\em Proceedings of the IEEE Conference on Computer Vision and
  Pattern Recognition}, pages 4733--4742, 2016.

\bibitem{chen20163d}
C.-H. Chen and D.~Ramanan.
\newblock 3d human pose estimation= 2d pose estimation+ matching.
\newblock {\em arXiv preprint arXiv:1612.06524}, 2016.

\bibitem{chen2016deeplab}
L.-C. Chen, G.~Papandreou, I.~Kokkinos, K.~Murphy, and A.~L. Yuille.
\newblock Deeplab: Semantic image segmentation with deep convolutional nets,
  atrous convolution, and fully connected crfs.
\newblock {\em arXiv preprint arXiv:1606.00915}, 2016.

\bibitem{cho2015accurate}
E.~Cho and D.~Kim.
\newblock Accurate human pose estimation by aggregating multiple pose
  hypotheses using modified kernel density approximation.
\newblock {\em IEEE Signal Processing Letters}, 22(4):445--449, 2015.

\bibitem{chu2016structured}
X.~Chu, W.~Ouyang, H.~Li, and X.~Wang.
\newblock Structured feature learning for pose estimation.
\newblock In {\em Proceedings of the IEEE Conference on Computer Vision and
  Pattern Recognition}, pages 4715--4723, 2016.

\bibitem{chu2017multi}
X.~Chu, W.~Yang, W.~Ouyang, C.~Ma, A.~L. Yuille, and X.~Wang.
\newblock Multi-context attention for human pose estimation.
\newblock {\em arXiv preprint arXiv:1702.07432}, 2017.

\bibitem{dong2014towards}
J.~Dong, Q.~Chen, X.~Shen, J.~Yang, and S.~Yan.
\newblock Towards unified human parsing and pose estimation.
\newblock In {\em Proceedings of the IEEE Conference on Computer Vision and
  Pattern Recognition}, pages 843--850, 2014.

\bibitem{fan2014pose}
X.~Fan, K.~Zheng, Y.~Zhou, and S.~Wang.
\newblock Pose locality constrained representation for 3d human pose
  reconstruction.
\newblock In {\em European Conference on Computer Vision}, pages 174--188.
  Springer, 2014.

\bibitem{gkioxari2016chained}
G.~Gkioxari, A.~Toshev, and N.~Jaitly.
\newblock Chained predictions using convolutional neural networks.
\newblock In {\em European Conference on Computer Vision}, pages 728--743.
  Springer, 2016.

\bibitem{gong2016human}
W.~Gong, X.~Zhang, J.~Gonz{\`a}lez, A.~Sobral, T.~Bouwmans, C.~Tu, and E.-h.
  Zahzah.
\newblock Human pose estimation from monocular images: A comprehensive survey.
\newblock {\em Sensors}, 16(12):1966, 2016.

\bibitem{he2016deep}
K.~He, X.~Zhang, S.~Ren, and J.~Sun.
\newblock Deep residual learning for image recognition.
\newblock In {\em Proceedings of the IEEE conference on computer vision and
  pattern recognition}, pages 770--778, 2016.

\bibitem{ionescu2014human3}
C.~Ionescu, D.~Papava, V.~Olaru, and C.~Sminchisescu.
\newblock Human3. 6m: Large scale datasets and predictive methods for 3d human
  sensing in natural environments.
\newblock {\em IEEE transactions on pattern analysis and machine intelligence},
  36(7):1325--1339, 2014.

\bibitem{jahangiri2017generating}
E.~Jahangiri and A.~L. Yuille.
\newblock Generating multiple diverse hypotheses for human 3d pose consistent
  with 2d joint detections.
\newblock In {\em Proceedings of the IEEE Conference on Computer Vision and
  Pattern Recognition}, pages 805--814, 2017.

\bibitem{jia2014caffe}
Y.~Jia, E.~Shelhamer, J.~Donahue, S.~Karayev, J.~Long, R.~Girshick,
  S.~Guadarrama, and T.~Darrell.
\newblock Caffe: Convolutional architecture for fast feature embedding.
\newblock In {\em Proceedings of the 22nd ACM international conference on
  Multimedia}, pages 675--678. ACM, 2014.

\bibitem{kohli2008simultaneous}
P.~Kohli, J.~Rihan, M.~Bray, and P.~H. Torr.
\newblock Simultaneous segmentation and pose estimation of humans using dynamic
  graph cuts.
\newblock {\em International Journal of Computer Vision}, 79(3):285--298, 2008.

\bibitem{ladicky2013human}
L.~Ladicky, P.~H. Torr, and A.~Zisserman.
\newblock Human pose estimation using a joint pixel-wise and part-wise
  formulation.
\newblock In {\em proceedings of the IEEE Conference on Computer Vision and
  Pattern Recognition}, pages 3578--3585, 2013.

\bibitem{li20143d}
S.~Li and A.~B. Chan.
\newblock 3d human pose estimation from monocular images with deep
  convolutional neural network.
\newblock In {\em Asian Conference on Computer Vision}, pages 332--347.
  Springer, 2014.

\bibitem{li2014heterogeneous}
S.~Li, Z.-Q. Liu, and A.~B. Chan.
\newblock Heterogeneous multi-task learning for human pose estimation with deep
  convolutional neural network.
\newblock In {\em Proceedings of the IEEE Conference on Computer Vision and
  Pattern Recognition Workshops}, pages 482--489, 2014.

\bibitem{li2015maximum}
S.~Li, W.~Zhang, and A.~B. Chan.
\newblock Maximum-margin structured learning with deep networks for 3d human
  pose estimation.
\newblock In {\em Proceedings of the IEEE International Conference on Computer
  Vision}, pages 2848--2856, 2015.

\bibitem{li2016fully}
Y.~Li, H.~Qi, J.~Dai, X.~Ji, and Y.~Wei.
\newblock Fully convolutional instance-aware semantic segmentation.
\newblock {\em arXiv preprint arXiv:1611.07709}, 2016.

\bibitem{liang2015human}
X.~Liang, C.~Xu, X.~Shen, J.~Yang, S.~Liu, J.~Tang, L.~Lin, and S.~Yan.
\newblock Human parsing with contextualized convolutional neural network.
\newblock In {\em Proceedings of the IEEE International Conference on Computer
  Vision}, pages 1386--1394, 2015.

\bibitem{lifshitz2016human}
I.~Lifshitz, E.~Fetaya, and S.~Ullman.
\newblock Human pose estimation using deep consensus voting.
\newblock In {\em European Conference on Computer Vision}, pages 246--260.
  Springer, 2016.

\bibitem{lin2016refinenet}
G.~Lin, A.~Milan, C.~Shen, and I.~Reid.
\newblock Refinenet: Multi-path refinement networks with identity mappings for
  high-resolution semantic segmentation.
\newblock {\em arXiv preprint arXiv:1611.06612}, 2016.

\bibitem{long2015fully}
J.~Long, E.~Shelhamer, and T.~Darrell.
\newblock Fully convolutional networks for semantic segmentation.
\newblock In {\em Proceedings of the IEEE Conference on Computer Vision and
  Pattern Recognition}, pages 3431--3440, 2015.

\bibitem{luvizon2017human}
D.~C. Luvizon, H.~Tabia, and D.~Picard.
\newblock Human pose regression by combining indirect part detection and
  contextual information.
\newblock {\em arXiv preprint arXiv:1710.02322}, 2017.

\bibitem{martinez2017simple}
J.~Martinez, R.~Hossain, J.~Romero, and J.~J. Little.
\newblock A simple yet effective baseline for 3d human pose estimation.
\newblock {\em arXiv preprint arXiv:1705.03098}, 2017.

\bibitem{mehta2016monocular}
D.~Mehta, H.~Rhodin, D.~Casas, O.~Sotnychenko, W.~Xu, and C.~Theobalt.
\newblock Monocular 3d human pose estimation using transfer learning and
  improved cnn supervision.
\newblock {\em arXiv preprint arXiv:1611.09813}, 2016.

\bibitem{moreno20163d}
F.~Moreno-Noguer.
\newblock 3d human pose estimation from a single image via distance matrix
  regression.
\newblock {\em arXiv preprint arXiv:1611.09010}, 2016.

\bibitem{newell2016stacked}
A.~Newell, K.~Yang, and J.~Deng.
\newblock Stacked hourglass networks for human pose estimation.
\newblock In {\em European Conference on Computer Vision}, pages 483--499.
  Springer, 2016.

\bibitem{oliveira2016deep}
G.~L. Oliveira, A.~Valada, C.~Bollen, W.~Burgard, and T.~Brox.
\newblock Deep learning for human part discovery in images.
\newblock In {\em Robotics and Automation (ICRA), 2016 IEEE International
  Conference on}, pages 1634--1641. IEEE, 2016.

\bibitem{park20163d}
S.~Park, J.~Hwang, and N.~Kwak.
\newblock 3d human pose estimation using convolutional neural networks with 2d
  pose information.
\newblock In {\em Computer Vision--ECCV 2016 Workshops}, pages 156--169.
  Springer, 2016.

\bibitem{ramakrishna2012reconstructing}
V.~Ramakrishna, T.~Kanade, and Y.~Sheikh.
\newblock Reconstructing 3d human pose from 2d image landmarks.
\newblock {\em Computer Vision--ECCV 2012}, pages 573--586, 2012.

\bibitem{rogez2017lcr}
G.~Rogez, P.~Weinzaepfel, and C.~Schmid.
\newblock Lcr-net: Localization-classification-regression for human pose.
\newblock In {\em CVPR 2017-IEEE Conference on Computer Vision \& Pattern
  Recognition}, 2017.

\bibitem{shen2016object}
W.~Shen, K.~Zhao, Y.~Jiang, Y.~Wang, Z.~Zhang, and X.~Bai.
\newblock Object skeleton extraction in natural images by fusing
  scale-associated deep side outputs.
\newblock In {\em Proceedings of the IEEE Conference on Computer Vision and
  Pattern Recognition}, pages 222--230, 2016.

\bibitem{shotton2013real}
J.~Shotton, T.~Sharp, A.~Kipman, A.~Fitzgibbon, M.~Finocchio, A.~Blake,
  M.~Cook, and R.~Moore.
\newblock Real-time human pose recognition in parts from single depth images.
\newblock {\em Communications of the ACM}, 56(1):116--124, 2013.

\bibitem{sun2017human}
K.~Sun, C.~Lan, J.~Xing, W.~Zeng, D.~Liu, and J.~Wang.
\newblock Human pose estimation using global and local normalization.
\newblock {\em arXiv preprint arXiv:1709.07220}, 2017.

\bibitem{sun2017compositional}
X.~Sun, J.~Shang, S.~Liang, and Y.~Wei.
\newblock Compositional human pose regression.
\newblock {\em arXiv preprint arXiv:1704.00159}, 2017.

\bibitem{tekin2016structured}
B.~Tekin, I.~Katircioglu, M.~Salzmann, V.~Lepetit, and P.~Fua.
\newblock Structured prediction of 3d human pose with deep neural networks.
\newblock {\em arXiv preprint arXiv:1605.05180}, 2016.

\bibitem{tekin2017learning}
B.~Tekin, P.~Marquez~Neila, M.~Salzmann, and P.~Fua.
\newblock Learning to fuse 2d and 3d image cues for monocular body pose
  estimation.
\newblock In {\em International Conference on Computer Vision (ICCV)}, number
  EPFL-CONF-230311, 2017.

\bibitem{tekin2016direct}
B.~Tekin, A.~Rozantsev, V.~Lepetit, and P.~Fua.
\newblock Direct prediction of 3d body poses from motion compensated sequences.
\newblock In {\em Proceedings of the IEEE Conference on Computer Vision and
  Pattern Recognition}, pages 991--1000, 2016.

\bibitem{tome2017lifting}
D.~Tome, C.~Russell, and L.~Agapito.
\newblock Lifting from the deep: Convolutional 3d pose estimation from a single
  image.
\newblock {\em arXiv preprint arXiv:1701.00295}, 2017.

\bibitem{tompson2014joint}
J.~J. Tompson, A.~Jain, Y.~LeCun, and C.~Bregler.
\newblock Joint training of a convolutional network and a graphical model for
  human pose estimation.
\newblock In {\em Advances in neural information processing systems}, pages
  1799--1807, 2014.

\bibitem{tripathi2017pose2instance}
S.~Tripathi, M.~Collins, M.~Brown, and S.~Belongie.
\newblock Pose2instance: Harnessing keypoints for person instance segmentation.
\newblock {\em arXiv preprint arXiv:1704.01152}, 2017.

\bibitem{wei2016convolutional}
S.-E. Wei, V.~Ramakrishna, T.~Kanade, and Y.~Sheikh.
\newblock Convolutional pose machines.
\newblock In {\em Proceedings of the IEEE Conference on Computer Vision and
  Pattern Recognition}, pages 4724--4732, 2016.

\bibitem{wu2016single}
J.~Wu, T.~Xue, J.~J. Lim, Y.~Tian, J.~B. Tenenbaum, A.~Torralba, and W.~T.
  Freeman.
\newblock Single image 3d interpreter network.
\newblock In {\em European Conference on Computer Vision}, pages 365--382.
  Springer, 2016.

\bibitem{xia2015zoom}
F.~Xia, P.~Wang, L.-C. Chen, and A.~L. Yuille.
\newblock Zoom better to see clearer: Human part segmentation with auto zoom
  net.
\newblock {\em arXiv preprint arXiv:1511.06881}, 2015.

\bibitem{xia2017joint}
F.~Xia, P.~Wang, X.~Chen, and A.~Yuille.
\newblock Joint multi-person pose estimation and semantic part segmentation.
\newblock {\em arXiv preprint arXiv:1708.03383}, 2017.

\bibitem{xie2015holistically}
S.~Xie and Z.~Tu.
\newblock Holistically-nested edge detection.
\newblock In {\em Proceedings of the IEEE international conference on computer
  vision}, pages 1395--1403, 2015.

\bibitem{yamaguchi2012parsing}
K.~Yamaguchi, M.~H. Kiapour, L.~E. Ortiz, and T.~L. Berg.
\newblock Parsing clothing in fashion photographs.
\newblock In {\em Computer Vision and Pattern Recognition (CVPR), 2012 IEEE
  Conference on}, pages 3570--3577. IEEE, 2012.

\bibitem{yang2016end}
W.~Yang, W.~Ouyang, H.~Li, and X.~Wang.
\newblock End-to-end learning of deformable mixture of parts and deep
  convolutional neural networks for human pose estimation.
\newblock In {\em Proceedings of the IEEE Conference on Computer Vision and
  Pattern Recognition}, pages 3073--3082, 2016.

\bibitem{yasin2016dual}
H.~Yasin, U.~Iqbal, B.~Kruger, A.~Weber, and J.~Gall.
\newblock A dual-source approach for 3d pose estimation from a single image.
\newblock In {\em Proceedings of the IEEE Conference on Computer Vision and
  Pattern Recognition}, pages 4948--4956, 2016.

\bibitem{zhou2017towards}
X.~Zhou, Q.~Huang, X.~Sun, X.~Xue, and Y.~Wei.
\newblock Towards 3d human pose estimation in the wild: A weakly-supervised
  approach.
\newblock In {\em Proceedings of the IEEE Conference on Computer Vision and
  Pattern Recognition}, pages 398--407, 2017.

\bibitem{zhou2016deep}
X.~Zhou, X.~Sun, W.~Zhang, S.~Liang, and Y.~Wei.
\newblock Deep kinematic pose regression.
\newblock In {\em Computer Vision--ECCV 2016 Workshops}, pages 186--201.
  Springer, 2016.

\bibitem{zhou2017sparse}
X.~Zhou, M.~Zhu, S.~Leonardos, and K.~Daniilidis.
\newblock Sparse representation for 3d shape estimation: A convex relaxation
  approach.
\newblock {\em IEEE transactions on pattern analysis and machine intelligence},
  39(8):1648--1661, 2017.

\bibitem{zhou2016sparseness}
X.~Zhou, M.~Zhu, S.~Leonardos, K.~G. Derpanis, and K.~Daniilidis.
\newblock Sparseness meets deepness: 3d human pose estimation from monocular
  video.
\newblock In {\em Proceedings of the IEEE Conference on Computer Vision and
  Pattern Recognition}, pages 4966--4975, 2016.

\end{thebibliography}
}

\end{document}